\documentclass[runningheads]{llncs}

\usepackage{eccv}

\usepackage{eccvabbrv}

\usepackage{graphicx}
\usepackage{booktabs}
\usepackage{multicol}
\usepackage{multirow}
\usepackage[accsupp]{axessibility}  % Improves PDF readability for those with disabilities.

\usepackage{hyperref}

\usepackage{orcidlink}

\begin{document}

\title{A Multimodal Benchmark Dataset and Model for Crop Disease Diagnosis} 


\author{Xiang Liu$\dag$\inst{1,2}\orcidlink{0009-0003-2492-403X} \and
Zhaoxiang Liu$\dag$*\inst{1,2}\orcidlink{0000-0002-1267-0277} \and
Huan Hu\inst{1,2}\orcidlink{0009-0009-6697-3220} \and
Zezhou Chen\inst{1,2}\orcidlink{0009-0000-6796-6043} \and
Kohou Wang\inst{1,2}\orcidlink{0009-0007-5863-2288} \and
Kai Wang\inst{1,2}\orcidlink{0000-0002-1171-0281} \and
Shiguo Lian*\inst{1,2}\orcidlink{0000-0003-4308-7049}} 

\authorrunning{X. Liu et al.}

\institute{AI Innovation Center, China Unicom, Beijing 100013, China \and
Unicom Digital Technology, China Unicom, Beijing 100013, China \email
{\{liux750,liuzx178,huh30,chenzz51,wangzp103,wangk115,liansg\}@chinaunicom.cn 
$\dag$Equal contribution,
*Corresponding author(s)}}
% \\
% \url{http://www.springer.com/gp/computer-science/lncs} 

\maketitle

\begin{abstract}
  While conversational generative AI has shown considerable potential in enhancing decision-making for agricultural professionals, its exploration has predominantly been anchored in text-based interactions. The evolution of multimodal conversational AI, leveraging vast amounts of image-text data from diverse sources, marks a significant stride forward. However, the application of such advanced vision-language models in the agricultural domain, particularly for crop disease diagnosis, remains underexplored. In this work, we present the crop disease domain multimodal (CDDM) dataset, a pioneering resource designed to advance the field of agricultural research through the application of multimodal learning techniques. The dataset comprises 137,000 images of various crop diseases, accompanied by 1 million question-answer pairs that span a broad spectrum of agricultural knowledge, from disease identification to management practices. By integrating visual and textual data, CDDM facilitates the development of sophisticated question-answering systems capable of providing precise, useful advice to farmers and agricultural professionals. We demonstrate the utility of the dataset by finetuning state-of-the-art multimodal models, showcasing significant improvements in crop disease diagnosis. Specifically, we employed a novel finetuning strategy that utilizes low-rank adaptation (LoRA) to finetune the visual encoder, adapter and language model simultaneously. Our contributions include not only the dataset but also a finetuning strategy and a benchmark to stimulate further research in agricultural technology, aiming to bridge the gap between advanced AI techniques and practical agricultural applications. The dataset is available at \url{https://github.com/UnicomAI/UnicomBenchmark/tree/main/CDDMBench}. 
\end{abstract}

\section{Introduction}

The accurate diagnosis of crop diseases is pivotal in agricultural practices, offering invaluable support to individuals involved in farming activities. However, prevailing solutions for crop disease diagnosis predominantly rely on single-modal approaches, such as visual algorithms for classification \cite{khamparia2020seasonal,agarwal2020potato} and detection \cite{saleem2019plant}. These single-modal methods can only provide diagnosis results and cannot offer richer agricultural knowledge based on human preferences. The imperative for a multimodal visual question answering (VQA) system \cite{antol2015vqa,fukui2016multimodal}, proficient in diagnosing crop diseases while possessing extensive agricultural knowledge, becomes evident. Such a system, facilitated by multimodal interactions and multi-round conversations, has the potential to expedite the retrieval of information concerning specific diseases and recommend appropriate mitigation strategies.

\begin{figure}[tb]
  \centering 
  \includegraphics[width=\textwidth,height=\textheight,keepaspectratio]{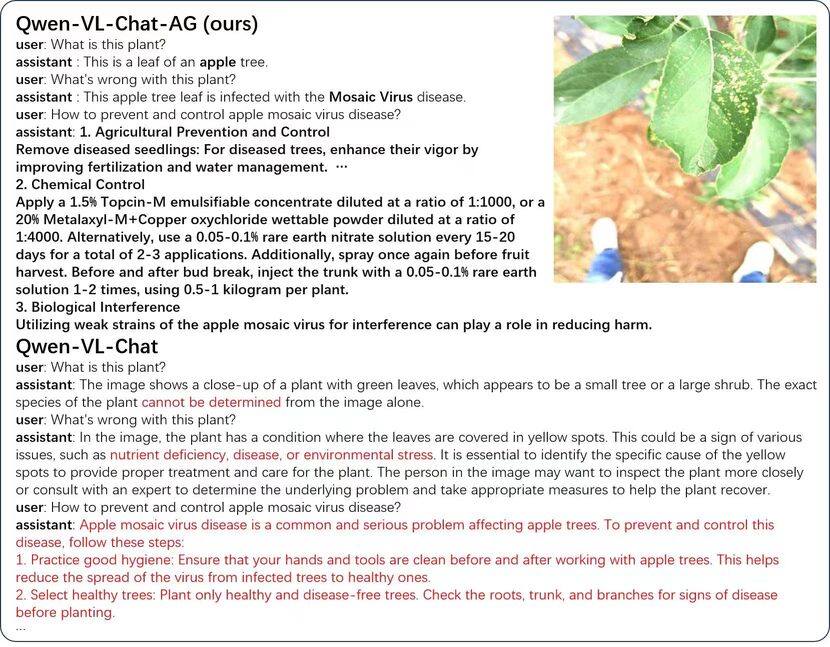}
  \caption{Example comparison of LVLMs on crop disease diagnosis. Our model accurately identifies crop and disease categories, offering detailed prevention and treatment methods. In contrast, Qwen-VL-Chat fails to determine both crop and disease categories, and provides detailed prevention and treatment methods, as indicated by the red texts.}
  \label{fig:figure1}
\end{figure}

\begin{figure}[tb]
  \centering
  \includegraphics[width=\textwidth,height=\textheight,keepaspectratio]{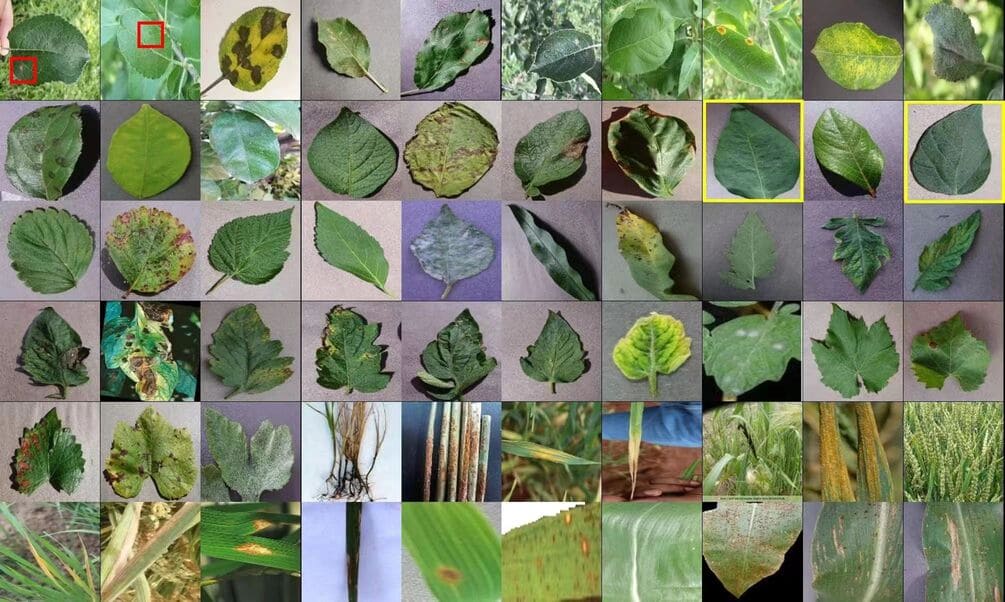}
  \caption{Examples of the crop disease image dataset. Each image represents a different category, and the leaves show a high degree of similarity, from their colors to their shapes. Additionally, some spot diseases display very similar visual features. Among the images, the two marked with red boxes represent different diseases but look very similar; the two marked with yellow boxes belong to different types of crops but have a very similar shape.}
  \label{fig:figure2}
\end{figure}
 
Most popular large-scale vision-language models (LVLMs), such as minigpt4 \cite{zhu2023minigpt}, Flamingo \cite{alayrac2022flamingo}, LLaVA \cite{liu2024visual} and Qwen-VL \cite{bai2023qwen-vl}, while effective as general-purpose multimodal conversational assistants \cite{askell2021general,li2022elevater,gan2022vision}, struggle in the crop disease domain. For example, Qwen-VL struggles with accurately identifying crop types, diagnosing diseases, and falls short in providing detailed strategies for disease prevention and control, as shown in Figure \ref{fig:figure1}. Consequently, developing a multimodal dataset specifically for adapting LVLMs to the crop disease domain is essential to enhance their accuracy and utility in agriculture. Building on this need, our work introduces a CDDM dataset.

Current finetuning strategy for LVLMs involves freezing the visual encoder while adjusting the projection/adapter module and the language model \cite{liu2024visual}. However, this approach faces challenges in the crop disease domain, due to the similarity of different crop diseases (as shown in Figure \ref{fig:figure2}). Employing this strategy means the visual encoder's ability to differentiate similar samples is limited, ultimately impacting the accuracy of disease diagnosis. 

The contributions of our paper can be summarized as follows:

\begin{itemize}
  \item The CDDM Dataset: We have meticulously curated a comprehensive dataset, comprising 137,000 images of crop diseases, and constructed a diverse set of 1 million question-answering instances. As illustrated in Figure \ref{fig:figure3}, for each diseased crop image, we've crafted conversations that encompass a range of information, including the crop and disease categories, detailed disease insights, and prevention and control strategies. This dataset serves as a fundamental resource for training models, which are capable of comprehending and addressing queries related to crop diseases.

  \item Model Finetuning Strategy: We adopted a novel finetuning strategy utilizing the LoRA \cite{hu2021lora}, for training Qwen-VL-Chat on the CDDM dataset without freezing the visual encoder. Our experiments have validated the effectiveness of this strategy, demonstrating its potential to significantly improve diagnostic accuracy in the agricultural domain.

  \item Open-Source Initiative: To promote research in the field of agricultural multimodal learning, we intend to release our dataset and the corresponding model codebase to the public. This initiative aims to foster collaboration and propel further advancements in the development of multimodal question-answering systems tailored for agricultural applications.
\end{itemize}

In conclusion, our work not only addresses the current limitations in crop disease diagnosis but also lays the groundwork for the development of specialized multimodal question-answering systems in the agricultural domain. Through the dissemination of our dataset and finetuned models, we aspire to pave the way for more effective and knowledgeable agricultural assistants that can significantly contribute to the enhancement of crop yield and overall farm management.

\section{Related Work}
\subsection{Related Work on Traditional Agricultural Diseases Diagnosis}

Traditional agriculture heavily relies on the observation and expertise of farmers and experts for the diagnosis and treatment of crop diseases. However, with the advent of smart agriculture, there is a growing trend towards utilizing computer vision methods to assist in disease diagnosis. Arya \cite{arya2019comparative} employed AlexNet \cite{krizhevsky2012imagenet} architectures to detect diseases in mango and potato leaves. Yang \cite{yang2021citrus} proposed a multimodal feature fusion network that combines RGB image networks with hyperspectral band extraction networks to improve the recognition accuracy of citrus Huanglongbing. Furthermore, Morbekar \cite{morbekar2020crop} utilized YOLO \cite{redmon2016you} for object detection to identify plant diseases, while Divyanth \cite{divyanth2023two} introduced a two-stage segmentation method using YOLO for plant segmentation and a disease segmentation model. These methods offer limited information, falling short in effectively aiding farmers with crop disease prevention and control.

\subsection{Multimodal Methods}
Multimodal question answering combines image and text information for answering queries. In a conventional VQA model, an image and a question are input, with visual and language encoders embedding them respectively. Following this, a cross-modal attention block utilizes the embeddings from both visual and language sources to understand the connections between the image and the question, which is crucial for making the final prediction. Lan \cite{lan2023visual} created a question-answer pairs dataset for fruit tree diseases, employing a co-attention architecture aligning ResNet \cite{he2016deep} and BERT \cite{devlin2018bert} features to achieve simple agricultural multimodal question-answering. However, this VQA model supports a limited range of crop diseases. Due to the limitations of its architecture, it has restricted generalization capabilities, offering only simplistic answers and lacking comprehensive agricultural knowledge.

The most popular LVLM typically combines a visual encoder and a large language model \cite{achiam2023gpt,anil2023palm,brown2020language,bai2023qwen,gao2023llama} via an adapter or a projector, where the former provides high-quality visual representation, and the latter offers powerful language generation capability. Li \cite{li2023blip} proposed using a lightweight querying transformer to align image encoders and frozen large language models. Liu \cite{liu2024visual} introduced LLaVA model which is trained with multimodal instruction-following data \cite{peng2023instruction}, and achieves a relative score of 85.1\% compared to GPT-4. Bai \cite{bai2023qwen-vl} presented the Qwen-VL model, which is trained on 1.14 billion image-text pairs. 
While these models excel in general domains, they are less effective in crop disease diagnosis because the image-text pairs specific to crop diseases differ significantly from the counterpart found on the web.

\section{Crop Disease Domain Multimodal Dataset}

\begin{figure}[tb]
  \centering
  \includegraphics[width=\textwidth,height=\textheight,keepaspectratio]{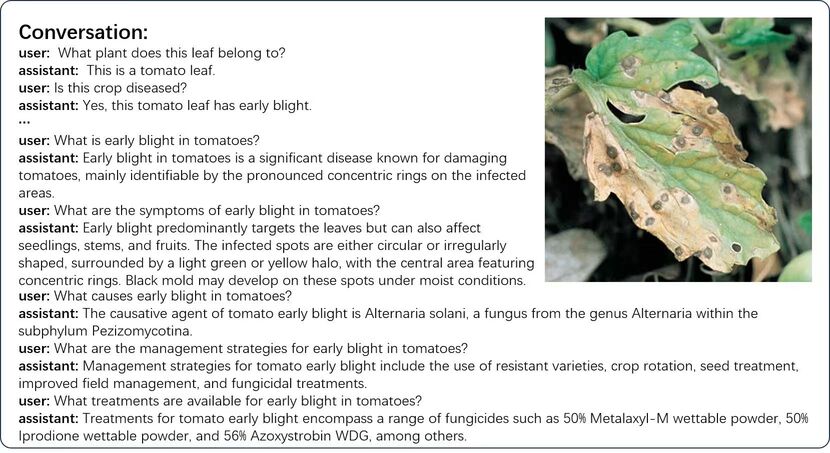}
  \caption{An instance of our CDDM data. The conversations cover the diagnosis, prevention, and treatment of crop diseases.}
  \label{fig:figure3}
\end{figure}

\begin{figure}[tb]
  \centering
  \includegraphics[width=\textwidth,height=\textheight,keepaspectratio]{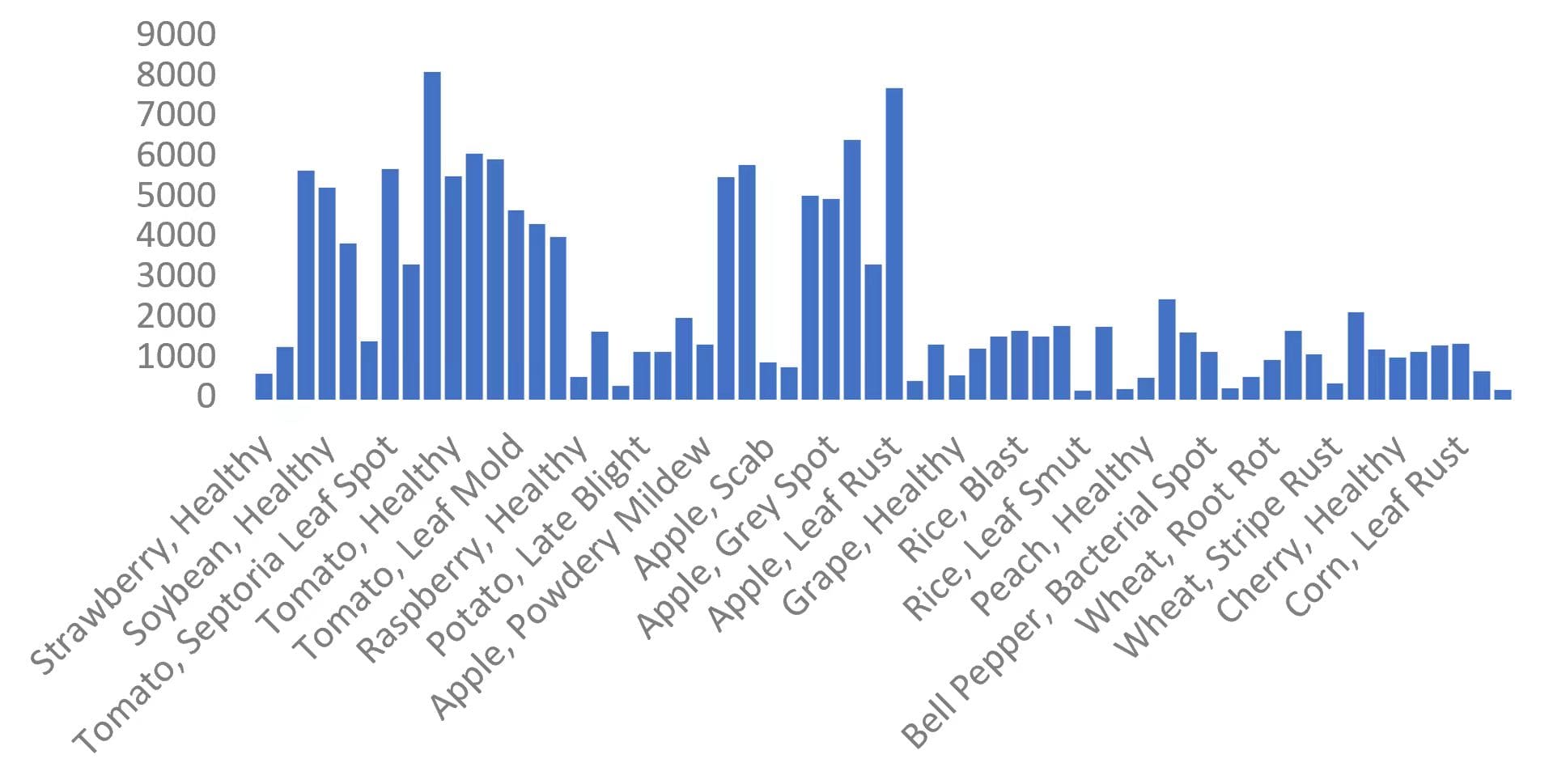}
  \caption{Distribution of the number of images for crop diseases dataset.}
  \label{fig:figure4}
\end{figure}

\begin{figure}[htbp]
  \centering
  \includegraphics[width=\textwidth,height=\textheight,keepaspectratio]{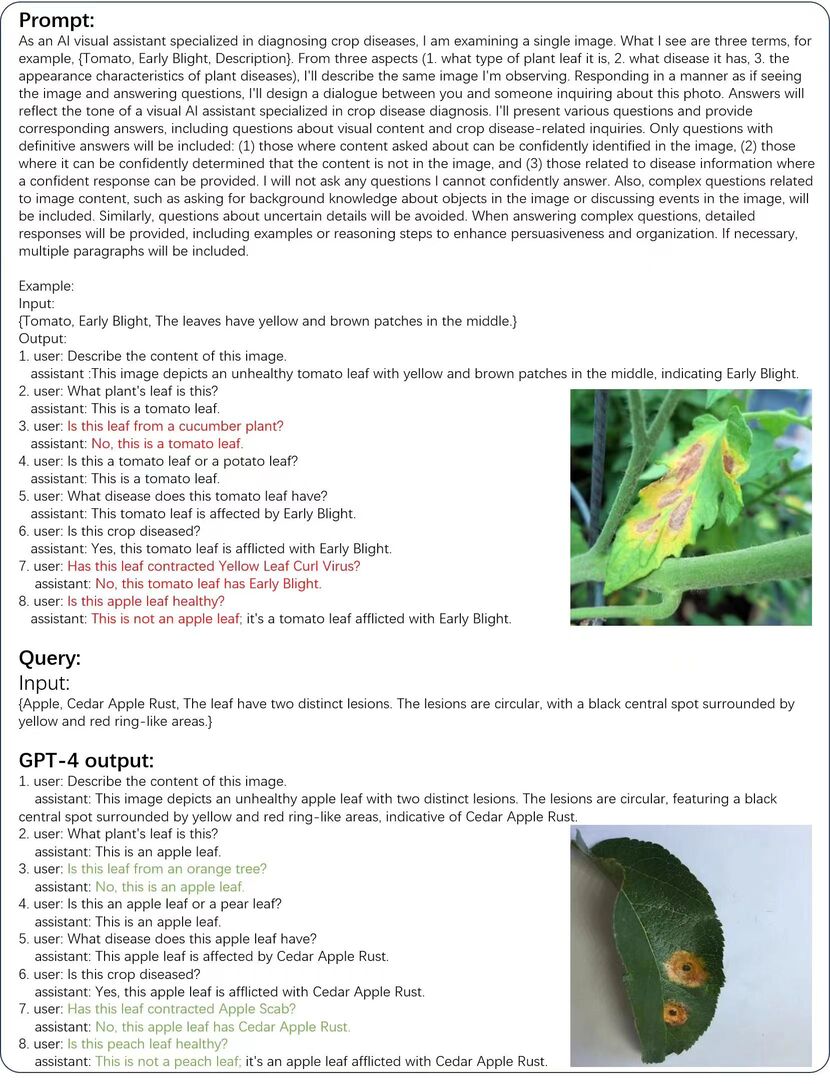}
  \caption{The prompt example of utilizing GPT-4 to generate instruction-following data of crop disease diagnosis. In the few-shot example within the "Prompt" part, the QA pairs highlighted in red are carefully crafted to include negative responses. After sequentially entering the "Prompt" part and the "Query" part, GPT-4 can generate 8 similar QA pairs, with negative responses highlighted in green.}
  \label{fig:figure5}
\end{figure}

To construct a robust crop disease diagnosis multimodal question-answering model, we developed a CDDM dataset. As depicted in Figure \ref{fig:figure3}, we present a sample of the CDDM data, comprising an image of crops alongside a series of interactive conversations. This dialogue encompasses a range of information, including the identification of the crop, diagnosis of any diseases affecting it, as well as details regarding the causes of these diseases and measures for their prevention and control. 

Here, we outline the construction process from three key aspects: crop image data collection and annotation, crop disease diagnosis instruction-following data, and crop disease knowledge instruction-following data.

\subsection{Crop Image Data Collection and Annotation}

The image data for this study consists of two parts: 

\begin{itemize}

    \item  Web Data: This includes agricultural datasets from Kaggle as well as agricultural disease data collected using web crawling methods, totaling 62,000 images. 
    \item  Private Data: We conducted field surveys in multiple farms and orchards to collect plant disease images, resulting in a total of 75,000 original images.

\end{itemize}

With the assistance of agricultural experts, we annotated all image data. The annotation information primarily includes crop category, disease category, and appearance description of the image.

Through data collection and annotation, we have compiled a dataset comprising 137,000 images. The dataset encompasses 16 categories of crops and a total of 60 categories of crop diseases. As illustrated in Figure \ref{fig:figure4}, there are 48 categories each containing more than 500 images, while 7 categories feature image counts ranging from 200 to 500. Overall, the distribution of data across categories is relatively even.

In addition, we collected text content of crop diseases knowledge related to crop diseases covering detailed descriptions including crop, disease, disease symptoms, pathogen characteristics, transmission pathways, disease conditions, prevention and control methods.

\subsection{Crop Disease Diagnosis Instruction-Following Data}
To enable the model to accurately diagnose crop diseases, we generate diverse instruction-following data through multi-round conversations about the provided crop images, utilizing language-only GPT-4 prompting. Specifically, we design few-shot instructions in a prompt that asks GPT-4 to generate questions and answers, with input \{crop category, disease category, appearance description\}. 

Experiments revealed that LVLMs tend to give affirmative responses more often in diagnosing plant species and disease categories. When posed with questions requiring negative answers, the visual model frequently errs by providing incorrect affirmative responses. Consequently, in crafting the question-answer corpus, we incorporated questions necessitating negative answers. 

% Figure \ref{fig:figure5} illustrates the prompt example of utilizing GPT-4 to generate instruction-following data for crop disease diagnosis. 

Figure \ref{fig:figure5} illustrates a prompt example for generating instruction-following data of crop disease diagnosis using GPT-4. In the "Prompt" part, we provide a few-shot example consisting of 8 questions and answers. The pairs highlighted in red are carefully crafted QA pairs with negative responses. After sequentially entering the content from the "Prompt" part and the "Query" part, GPT-4 can generate 8 similar QA pairs. The parts highlighted in green are the generated QA pairs with negative responses.

\begin{figure}[tb]
  \centering
  \includegraphics[width=\textwidth,height=\textheight,keepaspectratio]{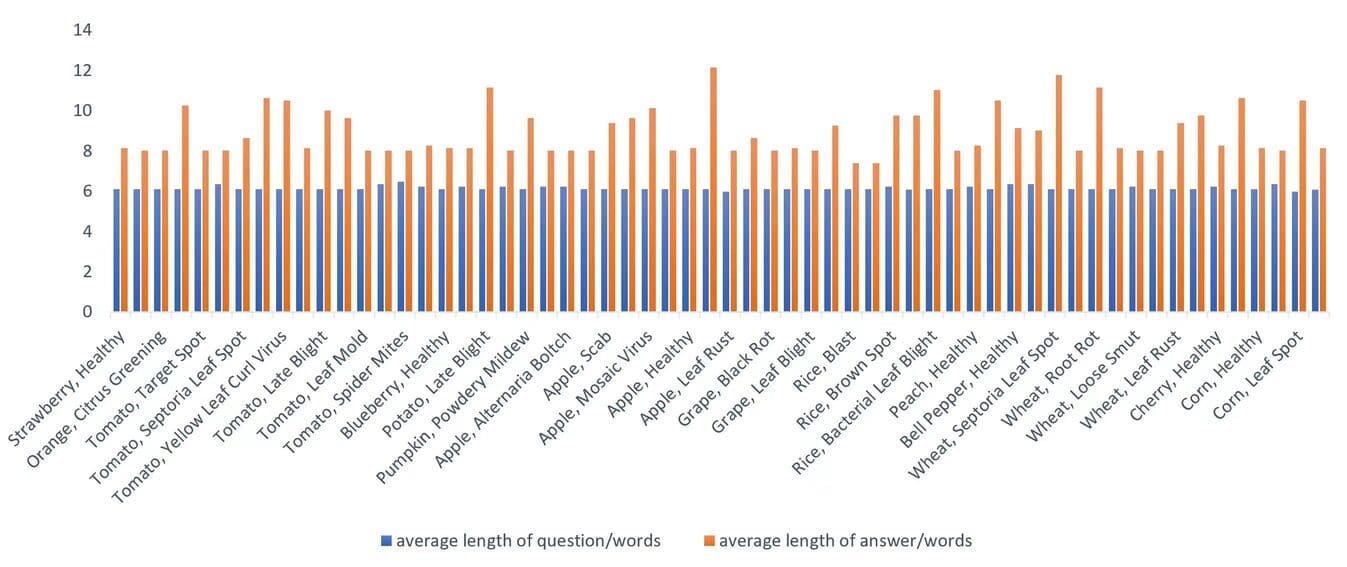}
  \caption{Distribution of the average lengths of questions and answers for different crop diseases in the CDDM Dataset.}
  \label{fig:figure6}
\end{figure}

The instruction-following dataset of crop disease diagnosis contains over 1 million question-answer pairs. The average question length is 6.11 words, while the average answer length is 8.92 words. Figure \ref{fig:figure6} shows the distribution of the average lengths of questions and answers for 60 types of crop diseases.

\subsection{Crop Disease Knowledge Instruction-Following Data}

\begin{figure}[tb]
  \centering
  \includegraphics[width=\textwidth,height=\textheight,keepaspectratio]{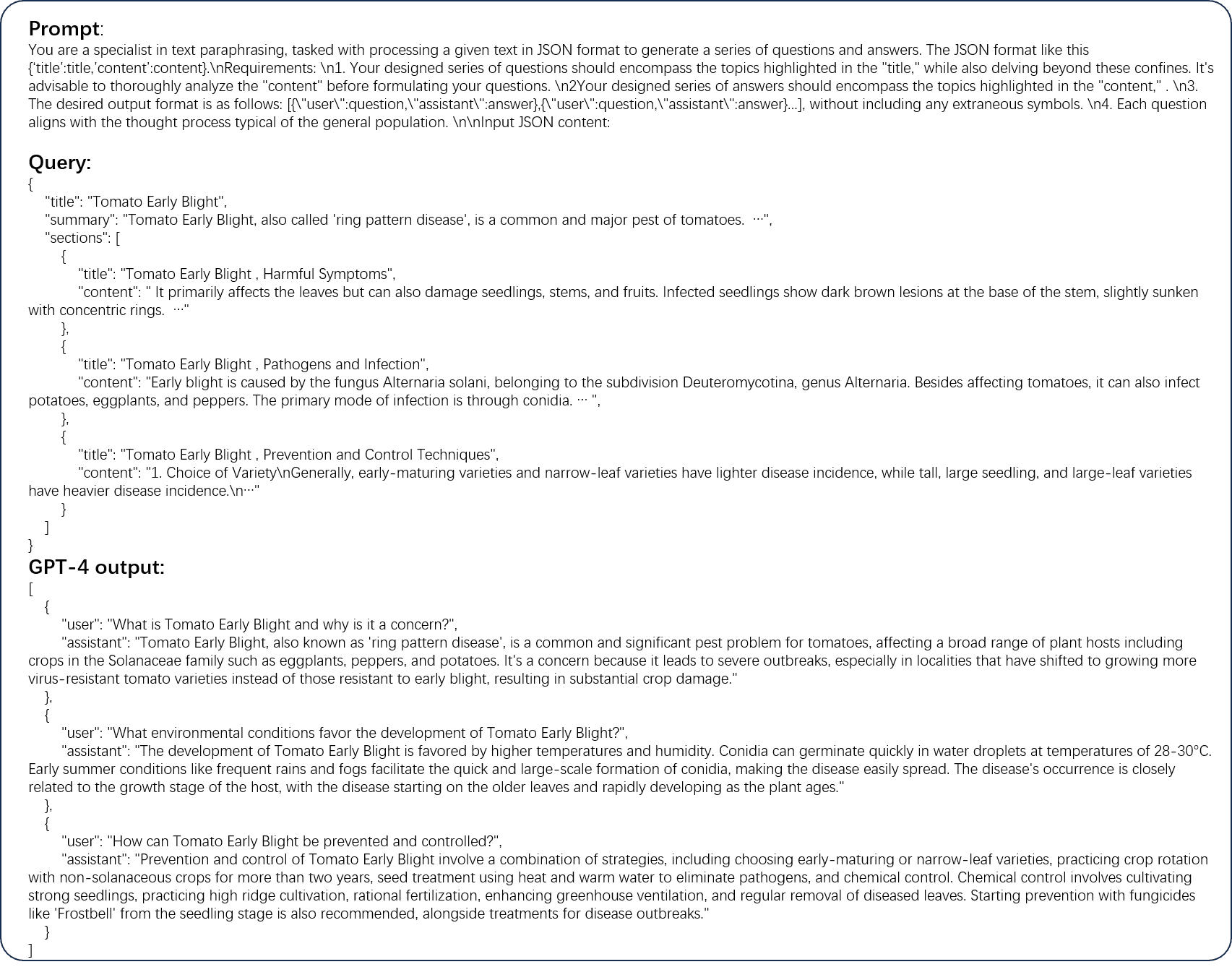}
  \caption{The prompt example of utilizing GPT-4 to generate instruction-following data of crop disease knowledge. After sequentially entering the "Prompt" section and the "Query" section, GPT-4 can generate high-quality QA pairs.}
  \label{fig:figure7}
\end{figure}

Based on the text content of crop disease knowledge we collected, we utilized GPT-4 to generate instruction-following conversations. The full prompt and the detailed example can be found in Figure \ref{fig:figure7}. Below is an example from the content in the "Query" section:

\textit{\textsl{\{"title": "Tomato Early Blight", "summary": "Tomato Early Blight, also called 'ring pattern disease', is a common and major pest of tomatoes. Some localities have been encouraged to grow more virus-resistant tomato varieties instead of those resistant to early blight, leading to severe blight outbreaks. The disease affects a broad range of plant hosts, including, but not limited to, crops in the Solanaceae family, such as eggplants, peppers, and potatoes."\}}}

Corresponding to this, the content of the question and answer pair generated by GPT-4 is presented as follows:

\textit{\textsl{\{"user": "What is Tomato Early Blight?","assistant": "Tomato Early Blight, also known as 'ring pattern disease', is a common and major pest of tomatoes that affects a broad range of plant hosts, including crops in the Solanaceae family such as eggplants, peppers, and potatoes."\}}}

In the crop disease knowledge instruction-following dataset, the average question length is 9.69 words, while the average answer length is 130.41 words.

{\bf The CDDM Dataset}: We constructed the CDDM dataset by merging crop disease diagnosis and knowledge instruction-following data and crop disease knowledge instruction-following data according to plant species and disease types.

\section{Adapting Multimodal Conversational Models to the Crop Disease Domain}

\begin{figure}[tb]
  \centering
  \includegraphics[width=6.5cm,keepaspectratio]{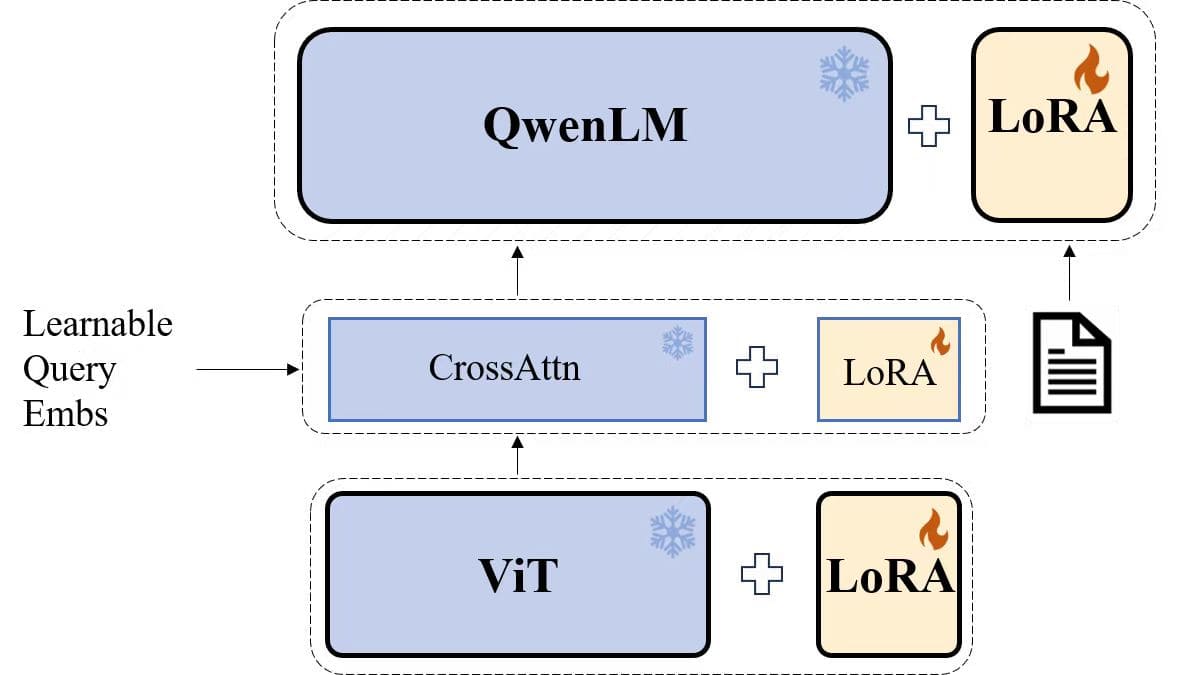}
  \caption{The LoRA training strategy on Qwen-VL-Chat.}
  \label{fig:figure8}
\end{figure}

We introduce a novel approach capable of adapting general LVLMs to agriculture focused LVLMs. Here, we use Qwen-VL-Chat as an example. Qwen-VL-Chat integrates three critical components: a language model, a visual encoder, and a position-aware vision-language adapter, i.e. cross attention module. As illustrated in Figure \ref{fig:figure8}, our method involves a specialized training regime, utilizing the LoRA technique, tailored specifically towards the domain of crop disease diagnosis. 

Our primary goal is adapting the Qwen-VL-Chat model to accurately diagnose crop diseases. To achieve this, We employ the LoRA training technique to simultaneously adjust the parameters of all three components of the model: the language model, the visual encoder, and the position-aware vision-language adapter. Our finetuning strategy is quite different from the finetuning strategy in LLaVA and Qwen-VL-Chat where the parameters of the visual encoder are not updated during finetuning. 

\section{Experiments}

To evaluate the utility of our proposed dataset and fine-tuning strategy, we fine-tuned the LLaVA model and the Qwen-VL-Chat model using two different strategies on our dataset.
One strategy is without freezing the visual encoder, the other is with freezing the visual encoder.

The model versions and hyper-parameters are below:

{\bf LLaVA-v1.5-7B} hyper-parameters: \{batch size: 128, learning rate: 2e-4, epochs: 5, maximum sequence length: 2048, weight decay: 0\} 

{\bf Qwen-VL-Chat-7B} hyper-parameters: \{batch size: 128, learning rate: 1e-5, epochs: 5, maximum sequence length: 2048, weight decay: 0.1\}

\subsection{Evaluation Metrics}

{\bf Crop Disease Diagnosis Performance}.  To assess the model's efficacy in diagnosing crop diseases, we constructed a test set using 3,000 images that were not included in the training set. The test set included a variety of questions and answers to evaluate the model's performance comprehensively.
The performance was measured based on the accuracy of the model's responses, specifically detecting the keywords of crop category and disease category in its answers.

{\bf Crop Disease Knowledge VQA Performance}.  Mirroring the evaluation methodology used by LLaVA and LLaVA-Med \cite{li2024llava}, we employed GPT-4 to assess the quality of model's generated responses to crop disease knowledge questions. We started by manually selecting original data pertaining to 10 types of crop diseases and crafting 20 questions. We randomly selected 20 images in the 10 types of crop diseases. Responses were then solicited from candidate models based on the images and questions provided. To provide an approximate theoretical upper bound, We create a reference prediction based on the question and the text content of crop disease knowledge, using the text-only GPT-4.

The responses from both the candidate models and the reference GPT-4 predictions were evaluated by GPT-4 for their helpfulness, relevance, and accuracy. Each response was scored on a scale from 1 to 10, with higher scores indicating superior overall performance. We calculated the total score for the model across all questions, with the maximum possible score being 200. To normalize this to a 100-point scale, we converted the total score accordingly.

\begin{table}[htbp]
  % \centering
  \caption{Results on crop disease diagnosis and knowledge QA. * indicates freezing the visual encoder.}
  \resizebox{\textwidth}{!}{
  \begin{tabular}{cccc}
  \toprule
  \multirow{2}{*}{Model} & \multicolumn{2}{c}{Crop Disease Diagnosis} & \multirow{2}{*}{     Crop Disease Knowledge QA} \\
  \cline{2-3}
   & {Crop Classification     } & {          Disease Classification} &  \\
   \midrule
   Qwen-VL-Chat & 28.4\% & 5.0\% & 41 \\
   Qwen-VL-Chat-AG* & 84.4\% & 66.1\% & 88.5 \\
   Qwen-VL-Chat-AG & 97.4\% & 91.5\% & 84 \\
  \midrule
  LLaVA-v1.5-7b & 24.5\% & 5.9\% & 47.5 \\
  LLaVA-AG* & 94.3\% & 82.1\% & {\bf 98} \\
  LLaVA-AG & {\bf 98.0\%} & {\bf 91.8\%} & 96.5 \\
  \bottomrule
  \end{tabular}
  }
  \label{tab:table1}
  \end{table}

\subsection{Results}

Table \ref{tab:table1} shows the experimental results. As can be seen, the models finetuned on our dataset outperform the LLaVA model and Qwen-VL-Chat model with a large margin. Our dataset builds a connection between images and the concepts of crops and diseases. This connection helps the finetuned models align image features with the LLM word embedding and then improve the performances. The base models have poor accuracies because they are unable to fully model the connection without training on our dataset.

The models finetuned without the visual encoder frozen outperform those finetuned with the visual encoder frozen considerably. Due to the similarity in appearance of different crops and different diseases (as shown in Figure \ref{fig:figure2}), the frozen visual encoder falls short of capturing the local details and patterns that distinguish them because it is trained on the general domain dataset. Finetuning it on our dataset enhances its ability to capture these local details and patterns, which is why finetuning results in a significant performance jump. 

In Figure \ref{fig:figure1}, we showcase examples of dialogues between Qwen-VL-Chat-AG and Qwen-VL-Chat in the context of crop disease diagnosis. Qwen-VL-Chat-AG was able to precisely identify crop diseases and offer effective prevention and treatment solutions, demonstrating the significant value of the CDDM dataset and the finetuning strategy in developing professional agricultural chatbots.

\subsection{Limitations}
{\bf Handling diseases out of domain}: We conducted several tests and found that our fine-tuned models perform not well in handling diseases outside our dataset. We guess that in-context learning might be the potential solution to this issue, e.g., adding one/few examples out of domain in the prompt to guide the models handling diseases in and out of domains. And we leave it as our future work to explore.

\section{Conclusions}

We presented a CDDM dataset and a LoRA based finetuning strategy. A series of experiments are conducted to validate the utility of our dataset and the finetuning strategy.
The models trained on our dataset with the proposed finetuning strategy gain significantly in the performances of crop disease diagnosis and knowledge VQA. Our contributions include not only the dataset but also a finetuning strategy and a benchmark to stimulate further research in agricultural technology, aiming to bridge the gap between advanced AI techniques and practical agricultural applications.

% \clearpage  % TODO REVIEW/FINAL: This \clearpage needs to be removed from both review and camera-ready versions.

\bibliographystyle{splncs04}
\bibliography{main}
\end{document}